\documentclass{article}

\usepackage{times}
\usepackage{latexsym}
\usepackage{amssymb}
\usepackage{graphicx}
\usepackage{caption}
\usepackage{subcaption}

\usepackage{arxiv}

\usepackage[utf8]{inputenc} 
\usepackage[T1]{fontenc}    
\usepackage{hyperref}       
\usepackage{url}            
\usepackage{booktabs}       
\usepackage{amsfonts}       
\usepackage{nicefrac}       
\usepackage{microtype}      
\usepackage{lipsum}

\title{Citation Sentiment Changes Analysis}

\author{Haixia Liu}

\begin{document}
\maketitle

\begin{abstract}
Metrics for measuring the citation sentiment changes were introduced. Citation sentiment changes can be observed from global citation sentiment sequences (GCSSs). With respect to a cited paper, the citation sentiment sequences were analysed across a collection of citing papers ordered by the published time. For analysing GCSSs, Eddy Dissipation Rate (EDR) was adopted, with the hypothesis that the GCSSs pattern differences can be spotted by EDR based method. Preliminary evidence showed that EDR based method holds the potential for analysing a publication's impact in a time series fashion.
\end{abstract}

\keywords{Citation Sentiment Changes Analysis \and  Global Citation Sentiment Sequences \and  Eddy Dissipation Rate}
\section{Introduction}\label{sec:introduction}
Using natural language processing and machine learning techniques, researchers investigated the methodologies to classify the citation sentences into three categories: objective, positive and negative \cite{athar2011sentiment}. Collections of annotated corpus for the task of citation sentiment analysis (CSA) were created \footnote{\url{https://cl.awaisathar.com/citation-context-corpus/}\label{fnurl}}. Despite the effectiveness on the task of CSA, the above works can only capture the polarity of a single citation sentence towards the cited paper. To evaluate a citation’s impact on an idea, analyzing a single citation is not enough. Taking all the citations that cite the same paper in one publication may give an overall sentiment score of the cited paper, but evaluating the sentiment changes of the citing papers that have cited the same paper over time is also important. Because the changes may reflect the impact of the cited paper from fluctuation perspective. Fluctuation occurs when the citation sentiment is changed from positive to negative or from negative to positive. It is assumed that ideas with different fluctuation levels have different properties. For example, if the ideas are with strong fluctuation, it may indicate that the problem discussed in that paper is a hot topic, or the solution proposed in that paper needs to be improved. Thus, investigating citation sentiment changes may help researchers to find research gaps.

This study is aiming at answering the question: if the paper $P_C$ has
been cited by different papers $P_A$ and $P_B$, assuming $P_B$ was published earlier than $P_A$, how to model the citation sentiment changes (with respect to the paper
$P_C$ ) over time \footnote{ The length of time of the fluctuation is important. This is why when comparing global citation sentiment changes, only citation sequences covering relatively same year-span are considered.}? To tackle this question, Eddy Dissipation Rate (EDR) \cite{nastrom1997turbulence} and extra three metrics were applied for discovering the pattern differences.

The eddy dissipation rate (EDR) is a fundamental variable of the atmospheric circulation \cite{nastrom1997turbulence}, which is used to measure the turbulence intensity. The cube root of the EDR is the International Civil Aviation Organization standard turbulence reporting metric \cite{kim2020retrieval}. Using EDR as the metric of turbulence intensity was originally proposed by MacCready \cite{maccready1964standardization} and is operationally practical, since EDR is proportional to the root-mean-square (RMS) vertical acceleration experienced by an aircraft for specific flight conditions \cite{maccready1964standardization,cornman1995real, sharman2014description}. More specifically, EDR is a measure of the viscous diffusion of turbulent kinetic energy, which is converted from large scales to small ones within the inertial sub-range via eddies, where dominate \cite{frisch1995turbulence} is affected by viscous. EDR is also an important parameter in large eddy simulations in that it is adopted in the momentum equation. In order to calculate EDR from velocity or temperature fluctuations, other atmospheric variables that are indirectly influenced are estimated, such as the atmospheric refractive index \cite{thiermann1992measurement}. In this study, an analogy was made between the atmospheric circulation and the citation sentiment changes. There are similarities between turbulence intensity and the fluctuation level of citation sentiment changes. For example,
the citation sentiment degree can be analogized to the velocity or temperature index. The sequence of citation sentiment degrees is similar to the sequence of the air velocities, which are used for computing the EDR. Therefore, using EDR to measure citation sentiment changes is reasonable. The approach used in this study is directly inherited from the method introduced in the paper \cite{Nijhuis2016Assessment}, which was originated from the work \cite{Siebert2006Observations}. 

\section{Methods and Materials}\label{sec:definitions}

\subsection{Basic Concepts and Terms}
\paragraph{Citation Sentiment Changes}\label{ssec:sent_change}
Inspired by Nguyen et al. \cite{nguyen2012predicting}, the citation
sentiment changes between positive and negative are with higher priority, thus, the citations with neutral polarity are ignored. Citation sentiment change is defined in such a way that it counts for one citation sentiment change if the citation sentiment polarity is changed from positive to negative or from negative to positive. Examples reflecting the citation sentiment changes are demonstrated in Table ~\ref{democitationchangelocal}. Even though the paper W06-1615 was being cited by the same author in the same paper E09-3005, the sentiment is different when citing different parts of the paper. For example, the first 
row on the Table ~\ref{democitationchangelocal} is a positive citation, which stated that the Structural Correspondence Learning (SCL) \cite{plank2009structural} was used as a solution for solving Part-of-Speech tagging and sentiment analysis problems. The second row expressed the unclear conclusion after applying SCL to non-projective dependency parsing problem, which made the citation negative. The third row is a positive citation, because promising results were generated using SCL. These examples indicated that the citing paper tend to investigate several aspects of the same paper. Whether the citation is positive or not depends on the specific situation. It is important to observe the sentiment changes of the author towards the cited paper. 

 \begin{table*}
  \caption{\label{democitationchangelocal} Observations about citation sentiment changes - Internal }
 \centering
 \begin{tabular}{p{2.5cm}|p{2.5cm}|p{4.5cm}|p{2.5cm}}
   CitedP & CitingP & Citations & Polarity  \\
   \hline
   W06-1615	& E09-3005 & SCL has been applied successfully in NLP for Part-of-Speech tagging and Sentiment Analysis. & 1 \\
   W06-1615	& E09-3005 & An attempt was made in the CoNLL 2007 shared task to apply SCL to non-projective dependency parsing (Shimizu and Nakagawa, 2007), however, without any clear conclusions. & -1\\
   W06-1615	& E09-3005 & We report on our exploration of applying SCL to adapt a syntactic disambiguation model and show promising initial results. & 1 \\ 
 \end{tabular}
 \end{table*}

\paragraph{Internal Citation Sentiment Sequence (ICSS)}
Let  $X = \langle x_1, x_2, \dots, x_i, \dots, x_N \rangle$ be a sequence of sentences ($X$) that have cited the paper $P_C$ in the citing paper $P_A$. $X$ is represented by an ordered sequence of $N$ citation sentences\footnote{Ordered by the citation's position in the paper. The citation appeared earlier in the paper holds the lower index in the set $X$}. The ICSS $S_{l}$ is defined as follows: $S_{l} = ( sl_1, sl_2, \dots, sl_i, \dots, sl_N )$, where $ sl_i  \in  \left \{ 0, 1, -1 \right \} $   expressing the sentiment polarity $\left \{objective, positive, negative\right \}$ in the citation $x_i$. An example of ICSS is \textit{1 -1 1 1 -1}. More examples of ICSS with its computed sentiment ($sl_i$) are shown in Table ~\ref{changewithinonepaper1}. The citation sentiment was annotated manually by Athar \cite{athar2012detection}. They used a subset of the dataset from Athar \cite{athar2011sentiment}, which consists of 20 target papers. These 20 papers correspond to approximately 20\% of incoming citations in the original dataset. They contain a total of 1,555 citations from 854 citing papers. They used a four-class scheme for annotation. The sentences without any direct or indirect mention of the citation were excluded  from  the  context.   The  rest  of  the  sentences  were  labeled with  either  positive  (1),  negative  (-1) or neutral (0). Note that in this paper, the task is not about analyzing each citation sentence and therefore the results in this paper are not comparable to the ones in the papers \cite{athar2011sentiment,athar2012detection}. Instead, the citation sentence labels provided by Athar \cite{athar2012detection} were directly used to compute new metrics in this study. Two metrics utilizing ICSSs were proposed, namely, $countcitations_{l}$ and $pratio_{l}$, which are shown in equation ~\ref{eq:countcitations_{l}} and ~\ref{eq:pratio_{l}}. $countcitations_{l}$ was mainly used for calculating $countcitations_g$ and $pratio_{l}$ was used for deriving EDR. The reason using $pratio_{l}$ rather than $countcitations_{l}$ for computing EDR is that $pratio_{l}$ holds normalized values, which were tailored for EDR calculation.

\begin{table*}
\caption{\label{changewithinonepaper1}  Examples of citation sentiment sequences generated from one paper. }
\centering
\begin{tabular}{p{2.5cm}|p{2.5cm}|p{5.5cm}}
  CitedP & CitingP & Citation Sentiment flow  \\
  \hline
  P07-1033	& D08-1105 & 1 1 1 1 1 1 1 1  \\
  A92-1018	& W98-1110 & -1 -1 -1 -1 -1 -1 \\
  W02-1011	& P09-1028 & 1 -1 1 1 -1 \\
  J90-1003	& D08-1007 & 1 -1 -1 -1 1 -1 -1 1 -1 \\  
\end{tabular}
\end{table*}

\paragraph{Global Citation Sentiment Sequence (GCSS)}
Let $D$ be the documents that have cited the same paper $P_C$. $D = \langle d_1, d_2, \dots, d_i, \dots, d_N \rangle$, then the GCSS $S_g$ is defined as : $S_g = ( sg_1, sg_2, \dots, sg_i, \dots, sg_N )$, where $sg_i \in \mathbb{Z}$ expressing the overall sentiment degree in the citing paper $d_i$.



\subsection{Citation Sentiment Score Computation}\label{p:score}
\begin{itemize}
\item  Measurement IN-1 \footnote{IN indicates this measurement is to measure internal citations}: $countcitations_{l}$\footnote{$l$ indicates local} 

For a specific cited paper, let $N_{pos}$ and $N_{neg}$ be the number of positive and negative citations in a citing paper:
\begin{equation}
countcitations_{l} = N_{pos} - N_{neg} \label{eq:countcitations_{l}}
\end{equation}

\item  Measurement IN-2: $pratio_{l}$
\begin{equation}
pratio_{l} = N_{pos} / (N_{pos} + N_{neg}) \label{eq:pratio_{l}} 
\end{equation}

\item  Measurement G-1 \footnote{G indicates this measurement is to measure citations globally}: $countcitations_g$\footnote{$g$ indicates global} 
\begin{equation}
countcitations_g = \sum\limits_{i=1}^n {countcitations_{l}}_{i}\label{eq:countcitations_g}
\end{equation}

where $n$ represents the total number of papers that have cited the specific paper.

\item  Measurement G-2: $countpapers_g$ 

For a specific cited paper, let $M_{pos}$ and $M_{neg}$ be the total number of positive and negative citing papers:
\begin{equation}
countpapers_g = M_{pos} - M_{neg}\label{eq:countpapers_g}
\end{equation}


\item  Measurement G-3: $pratiopaper_g$ 
\begin{equation}
pratiopaper_g = M_{pos} / (M_{pos} + M_{neg})\label{eq:pratiopaper_g}
\end{equation}

\end{itemize}

\subsection{Citation Sentiment Changes Analysis}\label{sec:gcss}


The purpose of analyzing citation sentiment changes is to discover some useful statistics that cannot be obtained by counting the number of positive and negative citations. In this study, the measurement $pratio_{l}$ proposed in Section ~\ref{p:score} was further utilized for generating higher level features to discover interesting patterns.
 
\begin{figure*}[!bth]
 \centering
 \includegraphics[width=0.99\textwidth]{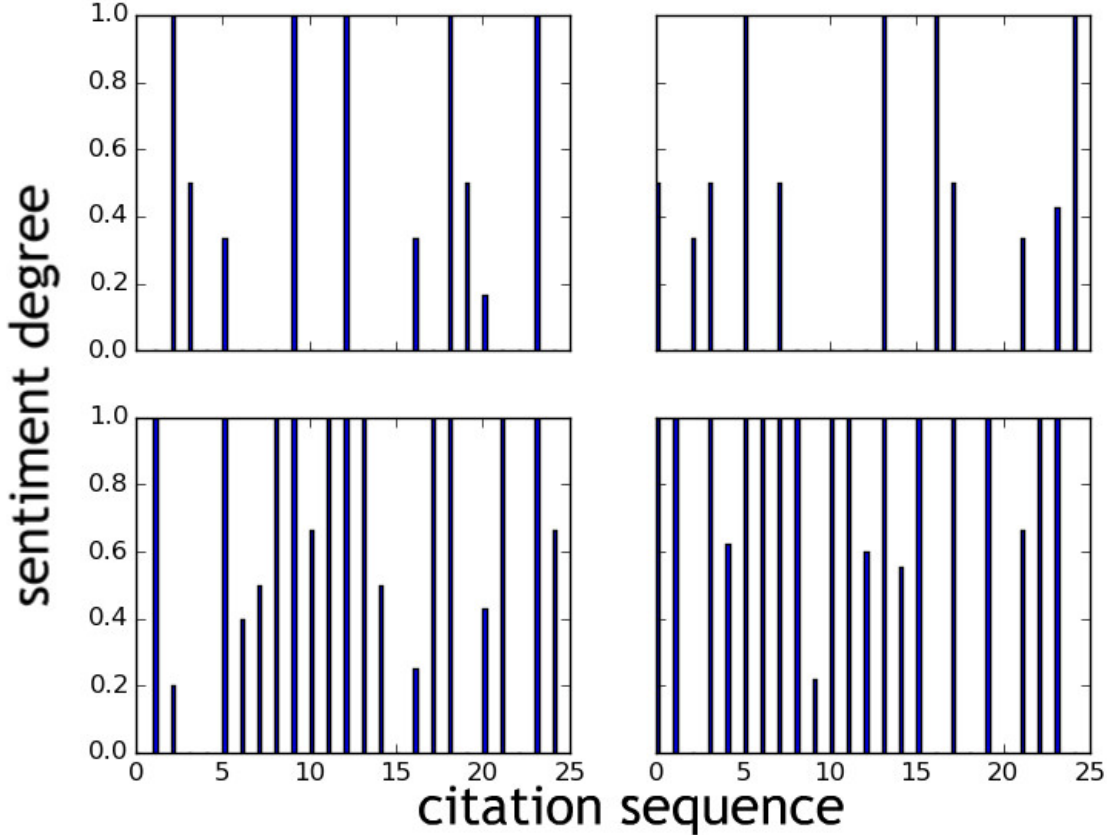}
 \vspace{-10pt}
 \caption{\label{fig:plot1234} Plots of four GCSSs. The horizontal-axis represents the length of the sequence and vertical-axis is the normalized sentiment value $pratio_{l}$. The range of the value is 0-1, where 1 indicates strong positive and 0 indicates strong negative. The place without bars means the $pratio_{l}$ value is 0. The bottom two sequence are subjectively controversial. Four sequences with detailed $pratio_{l}$ values are shown in Table ~\ref{datasetg}. The cited-paper IDs are: J93-1007 (top-left), J90-1003 (top-right), W05-0909 (bottom-left) and W04-1013 (bottom-right). }
 \end{figure*}

\paragraph{Manual Observations on The Citation Sentiment Sequences}
To manually analyze the citation sentiment changes over time, four plots of GCSSs are shown in Figure ~\ref{fig:plot1234} using the score $pratio_{l}$. The IDs shown in Figure ~\ref{fig:plot1234} can be retrieved from the annotated corpus, which are hosted by Athar et al. \cite{athar2012detection}. For each sub-figure, the horizontal-axis represents the number of citations with respect to a specific cited paper and the vertical-axis represents the normalized sentiment level, namely $pratio_{l}$. According the observations about the plots, the top two GCSSs in Figure ~\ref{fig:plot1234} are assumed to have less fluctuations \footnote{Note that the manual observations conducted in this study didn't take into account opinions from a substantial group of people. Therefore, bias may be introduced to the decisions about the fluctuation levels of these four GCSSs.} due to the reason that given the same length of the sequence, the top two bar-charts in Figure ~\ref{fig:plot1234} have more consecutive zeros. In contrast, there are more ups and downs in the bottom bar-charts. It is possible that papers with more fluctuations hold the potential for generating novel ideas since researchers can discover more research gaps in this kind of papers. On the contrary, papers having less fluctuations are either used as a solution for solving new problems or discarded when new approaches emerged.  

\paragraph{Eddy Dissipation Rate (EDR) for Measuring Fluctuation in GCSS}

Inspired by the methods for estimating turbulence intensity from water droplets \cite{Nijhuis2016Assessment}, their seven metrics were adopted as the citation sentiment turbulence indicator: non-periodic variance EDR, non-periodic power spectrum EDR, non-periodic 2nd order EDR, non-periodic 3rd order EDR, periodic power spectrum EDR, periodic 2nd order EDR and periodic 3rd order EDR. These seven metrics are the seven input features for the K-means cluster.

\subsection{Dataset for citation sentiment sequence analysis}

\begin{figure*}[!bth]
 \centering
 \includegraphics[width=1\textwidth]{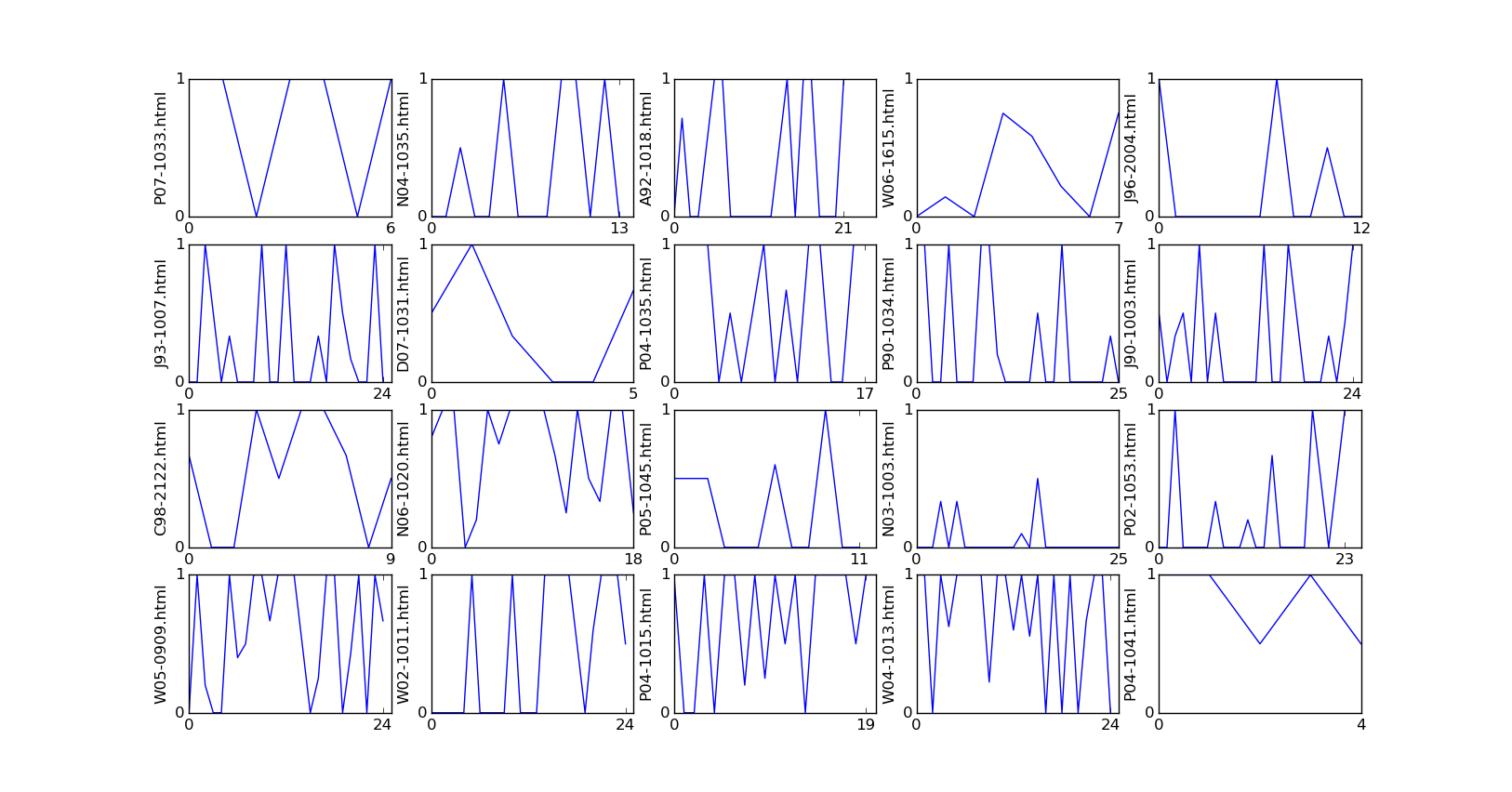}
 \vspace{-10pt}
 \caption{\label{fig:plotglobal20} Plots of 20 GCSSs. The horizontal-axis in the sub-figures represents the length of the citation sentiment sequence and vertical-axis is the normalized sentiment value $pratio_{l}$. The range of the value is [0,1], where 1 indicates strong positive and 0 strong negative. The label of each sub-figure can be found on Dr.Athar's website \cite{athar2012detection}.}
 \end{figure*}

In the paper \cite{athar2012detection}, the authors annotated 20 papers in a sentence-by-sentence manner (not annotating the paper as a complete unit). Based on these 20 annotated papers, global citation sentiment sequence dataset (DataG) was created. The DataG contains 20 $<$ cited-paper, citation-sentiment-sequence $>$ pairs, which were derived from 3581 $<$ cited-paper, citation-sentiment-sequence$>$ pairs (282 positive, 419 negative and 2880 neutral). The value in the citation-sentiment-sequence was obtained by analyzing the citation sentiment changes. Part of the DataG examples are shown in Table~\ref{datasetg} and the plots of the 20 GCSSs are shown in Figure ~\ref{fig:plotglobal20}\footnote{Using curves to represent the citation sentiment changes is to visualize the sequence turbulence. Some of them are also demonstrated in the bar-chart ~\ref{fig:plot1234}.}. Note that although each citation sentence has the ground truth label, the 20 papers do not have labels, which is the main reason that this study used unsupervised machine learning technique to analyze the data.

\begin{table*}
\caption{\label{datasetg}  Samples of DataG. The values in the citation sentiment sequence were computed using the proportion of positive citation counts among the total number of citations. The values are separated by space.}
\centering
\begin{tabular}{p{2cm}|p{7cm}}
  PaperID  &  Citation sentiment sequence\\
  \hline
  J93-1007	& 0.0 0.0 1.0 0.5 0.0 0.3 0.0 0.0 0.0 1.0 0.0 0.0 1.0 0.0 0.0 0.0 0.3 0.0 1.0 0.5 0.2 0.0 0.0 1.0 0.0 \\
  J90-1003	& 0.5 0.0 0.3 0.5 0.0 1.0 0.0 0.5 0.0 0.0 0.0 0.0 0.0 1.0 0.0 0.0 1.0 0.5 0.0 0.0 0.0 0.3 0.0 0.4 1.0\\
  W05-0909 & 0.0 1.0 0.2 0.0 0.0 1.0 0.4 0.5 1.0 1.0 0.7 1.0 1.0 1.0 0.5 0.0 0.25 1.0 1.0 0.0 0.4 1.0 0.0 1.0 0.7\\
  W04-1013 & 1.0 1.0 0.0 1.0 0.6 1.0 1.0 1.0 1.0 0.2 1.0 1.0 0.6 1.0 0.6 1.0 0.0 1.0 0.0 1.0 0.0 0.7 1.0 1.0 0.0\\
\end{tabular}
\end{table*}

\subsection{Machine Learning Algorithms}\label{subsec:ml}
Cluster method was used\footnote{\url{$http://scikit-learn.org/stable/modules/generated/sklearn.cluster.KMeans.html$}} to discover the different fluctuation patterns in the GCSSs. K-means algorithm \cite{krishna1999genetic} was adopted to cluster the 20 GCSSs. Two important factors affect the cluster results, which are the number of clusters and the input features. In this study, the number of clusters was determined by an evaluation method that's introduced in Section ~\ref{section:eval}. Three groups of features were used for clustering, one of which were computed by EDR method using $pratio_{l}$ shown in the equation ~\ref{eq:pratio_{l}}, namely, $pratio_{l}$ EDR based features. The dimension of $pratio_{l}$ EDR based features is seven, which are the seven metrics introduced in Section \ref{sec:gcss} \cite{Nijhuis2016Assessment}. Another group of features were based on the values $countcitations_g$, $countpapers_g$ and $pratio_g$, which were generated using the equations ~\ref{eq:countcitations_g}, ~\ref{eq:countpapers_g} and ~\ref{eq:pratiopaper_g} respectively, namely, $g_{value}$ based features. The dimension of $g_{value}$ based features is three. The third group of features were formed by concatenating the $pratio_{l}$ EDR based features and $g_{value}$ based features, which resulted in ten-dimension input features, namely, $concatenated$ features. These three groups of features were evaluated in Section ~\ref{section:eval}.

\section{Evaluation} \label{section:eval}
\begin{figure*}
     \centering
     \begin{subfigure}[b]{0.3\textwidth}
         \centering
         \includegraphics[width=\textwidth]{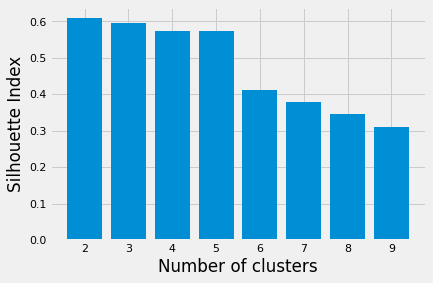}
         \caption{$pratio_{l}$ EDR based features}
         \label{fig:7edr}
     \end{subfigure}
     \hfill
     \begin{subfigure}[b]{0.3\textwidth}
         \centering
         \includegraphics[width=\textwidth]{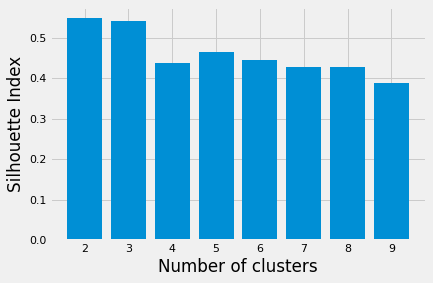}
         \caption{$g_{value}$ based features}
         \label{fig:3g}
     \end{subfigure}
     \hfill
     \begin{subfigure}[b]{0.3\textwidth}
         \centering
         \includegraphics[width=\textwidth]{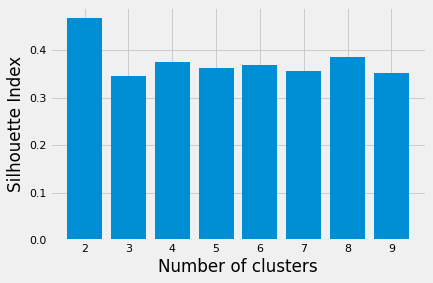}
         \caption{$concatenated$ features}
         \label{fig:concat}
     \end{subfigure}
        \caption{Average SI for each cluster. The average SIs for $pratio_{l}$ EDR (a), $g_{value}$ (b) and $concatenated$ (c) based features cluster methods are \textbf{0.61}, 0.55 and 0.47 respectively. The best SI was obtaining using $pratio_{l}$ EDR based features.}
        \label{fig:si}
\end{figure*}

Unlike supervised learning where the data has ground truth to analyze the method's performance, clustering is lacking of solid evaluation metrics for comparing different clustering strategies. One of the main purposes of cluster evaluation is to determine the optimal number of clusters. We can evaluate the performances of the methods based on different number of clusters \cite{maulik2002performance}. Compactness and separation are the two important measurement criteria for determining the optimal number of clusters \cite{berry2004data}. Compactness guarantees the samples in a cluster to be as close to each other as possible and the variance is the commonly used value for validating compactness. Separation ensures a cluster is well-separated from other clusters. Two widely accepted indices used for measuring separation are the distances between cluster centers and the pairwise minimum distances between samples in different clusters. The Silhouette Index (SI) was first introduced by Peter J. Rousseeuw in 1986 \cite{rousseeuw1987silhouettes}. It is used for interpreting and validating cluster data. The SI obtains the optimal clustering number by the difference between the average distance within the cluster and the minimum distance among different clusters. The  average  SI  gives  the  overall  clustering  quality  of  the  entire  data  set. If the average  SI  is close to 0, then it indicates the sample is very close to the neighboring clusters. If it is close to 1, then it means the sample is far away from the neighboring clusters. if it is close to -1, then it shows the sample is assigned to the wrong clusters. Therefore, in order to derive a good cluster, we want the average SI to be as big as possible and close to 1. In this study, SI was adopted for selecting the optimal number of clusters. Figure \ref{fig:si} illustrates a comparison of the average SI for each cluster. The number of clusters being evaluated is ranging from 2 to 9,  which is reasonable for testing a dataset with 20 samples. We can see that the highest average SI was achieved when the number of clusters was set to 2 for all the three types of features. Using $pratio_{l}$ EDR based features, the best average SI (0.61) was obtained. The average SIs for $g_{value}$ and $concatenated$ based features cluster methods are 0.55 and 0.47 respectively.

\section{Results}
The papers being clustered shown in Table ~\ref{tab:cluster} correspond to each other in the order shown in Figure ~\ref{fig:plotglobal20}. As we can see from Table ~\ref{tab:cluster} that using $pratio_{l}$ EDR based features, the papers P07-1033, W06-1615, D07-1031 and P04-1041 were assigned to the same cluster and the rest of the papers were assigned to the other cluster. When we look at Figure ~\ref{fig:plotglobal20}, papers P07-1033, W06-1615, D07-1031 and P04-1041 share similar characteristics in that they are less fluctuated and they have shorter sequences in comparison with the rest of the papers. On the one hand $pratio_{l}$ EDR based features were able to capture the GCSSs pattern differences, but on the other hand it seems the results were heavily affected by the length of the sequence. The cluster results derived by $concatenated$ based features are identical to the results generated by $pratio_{l}$ EDR based features. $g_{value}$ based features gave different cluster results in that it clustered the papers P04-1035, C98-2122, N06-1020, W05-0909, W02-1011, P04-1015 and W04-1013 to the opposite group. Despite the disagreement between $g_{value}$ based cluster and the other two methods, all the three methods group the papers P07-1033, W06-1615, D07-1031 and P04-1041 into the same cluster. The reason that the best SI was achieved by using EDR method maybe due to EDR's specialty on sequence data analysis. Because $g_{value}$ features were mainly accumulated values, they are better at spotting patterns from static perspective.

\begin{table*}
\caption{\label{tab:cluster} GCSSs clustering results using K-means. Each cell contains the cited paper ID and the cluster labels generated by $pratio_{l}$ EDR based features, $concatenated$ features and $g_{value}$ based features. The cluster labels were presented bellow each paper ID with the order ($pratio_{l}$ EDR based features / $concatenated$ features / $g_{value}$ based features). The 20 cells in this table were organized in an order correspond to the ones shown in Figure ~\ref{fig:plotglobal20}. Note that the cluster label 1 and 0 in this table do not indicate citation sentiment polarity and neither do they indicate fluctuation level. The labels simply indicate that the samples with the same label share similar patterns.}



 \centering
 \begin{tabular}{p{2.cm}|p{2.cm}|p{2.cm}|p{2.cm}|p{2.cm}}
 	
 P07-1033 & N04-1035 & A92-1018 & W06-1615 & J96-2004 \\ 
 (1/1/1) & (0/0/0) & (0/0/0) & (1/1/1) & (0/0/0) \\  
 \hline

 J93-1007 & D07-1031 & P04-1035 & P90-1034 & J90-1003 \\
 (0/0/0) & (1/1/1) & (0/0/1) & (0/0/0) & (0/0/0) \\
 \hline

 C98-2122 & N06-1020 & P05-1045 & N03-1003 & P02-1053 \\
 (0/0/1) & (0/0/1) & (0/0/0) & (0/0/0) & (0/0/0) \\
 \hline

 W05-0909 & W02-1011 & P04-1015 & W04-1013 & P04-1041 \\ 
 (0/0/1) & (0/0/1) & (0/0/1) & (0/0/1) & (1/1/1) \\   
   
 \end{tabular} 
 \end{table*}

\section{Conclusion and Discussion}
In this study, interesting approaches for measuring citation sentiment changes have been tested. Preliminary evidence showed that the proposed methods might be useful for measuring a publication’s impact in a time series fashion. The $pratio_{l}$ EDR based method for clustering GCSSs focused on evaluating the paper’s impact through time. Testing the citation sentiment changes over time is essentially testing the turbulence of a sequence. EDR method has been used in air turbulence evaluation, thus, it is chosen as the empirical method. For comparison, extra three metrics were introduced, which are $countcitations_g$, $countpapers_g$ and $pratio_g$. Unsupervised machine learning was carried out, aiming at discovering previously undetected patterns from the dataset with 20 unlabeled samples. K-means algorithm was used for clustering the data and average Silhouette Index (SI) was calculated for evaluating the cluster results. The average SI for each method indicated that $pratio_{l}$ EDR based features gave the best cluster result. Therefore, it could be concluded that $pratio_{l}$ EDR based K-means clustering approach is able to distinguish different patterns from GCSSs. However, due to lacking of annotated data, it is difficult to assign labels to the identified clusters with confidence. In order to overcome the data shortage, more GCSSs are needed. For example, 300 GCSSs would be good for further experiment.  More reliable results could be obtained using the data that are mixed with different level of turbulence samples. Different people have different opinions about the definition of a fluctuating Sequence. Therefore, collecting multiple labels for GCSSs from different people is necessary. The final label would be determined by the one that has most vote from the annotators.


One application of this work can be used as the scientometrics for personalized literature recommendation. According to different preferences from different people, they may selectively choose the papers with preferred fluctuation level to read. For example, for the purpose of discovering research gaps, papers with more fluctuated GCSSs maybe a good choice. 



\end{document}